\documentclass[11pt]{article}
\usepackage{coling2020}
\usepackage{times}
\usepackage{url}
\usepackage{latexsym}

\usepackage[hidelinks]{hyperref}
\usepackage{booktabs}
\usepackage{array}
\usepackage{makecell}
\usepackage{xspace}
\usepackage{amsthm}
\usepackage{multirow}
\usepackage{xargs}                      %
\usepackage{xcolor}  %
\usepackage{amsmath}
\usepackage{amsfonts}
\usepackage{dsfont}
\usepackage{listings}
\usepackage{bbm}
\usepackage{wrapfig}

\usepackage[colorinlistoftodos,prependcaption,textsize=tiny,disable]{todonotes}

\newcommand{\webnlg}{WebNLG\xspace}
\newcommand{\ldc}{LDC2017T10\xspace}
\newcommand{\viggo}{ViGGO\xspace}

\newcommand{\cleanedee}{Cleaned E2E\xspace}
\newcommand{\qdsa}{Q$_D$\xspace}
\newcommand{\qhsa}{Q$_H$\xspace}

\newcommand{\system}{\textsc{DataTuner}\xspace}

\newcommand{\systemFcPost}{\textsc{DataTuner\_fc}\xspace}

\newcommand{\systemNoFc}{\textsc{DataTuner\_no\_fc}\xspace}
\newcommand{\systemNoFcNoFs}{\textsc{DataTuner\_no\_fc/fs}\xspace}
\newcommand{\numData}{4\xspace}

\newcommand{\dtlm}{\textsc{D2T-LM}\xspace}

\newcolumntype{C}[1]{>{\centering\let\newline\\\arraybackslash\hspace{0pt}}m{#1}}

\newcommand{\sfc}{SFC\xspace}
\newcommand{\mr}{\textit{MR}\xspace}

\newtheoremstyle{break}
  {\topsep}{\topsep}%
  {\normalfont}{}%
  {\bfseries}{}%
  {\newline}{}%
\theoremstyle{break}

\theoremstyle{definition}
\theoremstyle{break}

\newcommandx{\unsure}[2][1=]{\todo[linecolor=red,backgroundcolor=red!25,bordercolor=red,#1]{#2}}
\newcommandx{\change}[2][1=]{\todo[linecolor=blue,backgroundcolor=blue!25,bordercolor=blue,#1]{#2}}
\newcommandx{\info}[2][1=]{\todo[linecolor=green,backgroundcolor=green!25,bordercolor=OliveGreen,#1]{#2}}
\newcommandx{\improvement}[2][1=]{\todo[linecolor=Plum,backgroundcolor=Plum!25,bordercolor=Plum,#1]{#2}}
\newcommandx{\thiswillnotshow}[2][1=]{\todo[disable,#1]{#2}}
\newcommand{\specialtoken}[1]{\mbox{\textless \textit{#1}\textgreater}}
\newcommand{\vertmulticell}[2]{\multirow{#1}{*}{\rotatebox[origin=c]{90}{#2}}}
\newcommand{\hormulticell}[2]{\multirow{#1}{*}{#2}}

\definecolor{codegreen}{rgb}{0,0.6,0}
\definecolor{codegray}{rgb}{0.5,0.5,0.5}
\definecolor{codepurple}{rgb}{0.58,0,0.82}
\definecolor{backcolour}{rgb}{0.95,0.95,0.92}
 
\lstdefinestyle{mystyle}{
    basicstyle=\footnotesize, %
    keywordstyle=\color{black}\bfseries,
    identifierstyle=, %
    commentstyle=\color{white}, %
    columns=fullflexible,
    morekeywords={D, Linearized,T, WebNLG, LDC2017T10, Cleaned, E2E, ViGGO},
    backgroundcolor=\color{backcolour}, 
    keywordsprefix={<},
}
\usepackage[font={normalsize}]{caption}

\colingfinalcopy %

\title{\textit{Have Your Text and Use It Too!}\\ End-to-End Neural Data-to-Text Generation with Semantic Fidelity}

\author{Hamza Harkous \\
  Amazon Alexa \\
  {\tt \small hamza.harkous@gmail.com} \\\And
  Isabel Groves \\
  Amazon Alexa \\
  {\tt \small  isabeg@amazon.com} \\ \And 
  Amir Saffari \\
  Amazon Alexa \\
  {\tt \small  amsafari@amazon.com} \\

  }
\date{}

\begin{document}

\maketitle

\begin{abstract}
End-to-end neural data-to-text (D2T) generation has recently emerged as an alternative to pipeline-based architectures. However, it has faced challenges generalizing to new domains and generating semantically consistent text.
In this work, we present \system, a neural, end-to-end data-to-text generation system that makes minimal assumptions about the data representation and target domain. We take a two-stage generation-reranking approach, combining a fine-tuned language model with a semantic fidelity classifier.
Each component is learnt end-to-end without needing dataset-specific heuristics, entity delexicalization, or post-processing. 
We show that \system achieves state of the art results on automated metrics across four major D2T datasets (\ldc, \webnlg, \viggo, and \cleanedee), with fluency assessed by human annotators as nearing or exceeding the human-written reference texts.
Our generated text has better semantic fidelity than the state of the art on these datasets.
We further demonstrate that our model-based semantic fidelity scorer %
is a better assessment tool compared to traditional heuristic-based measures of semantic accuracy. 
\end{abstract}

\section{Introduction}

\blfootnote{
    
    \hspace{-0.65cm}  %
    This work is licensed under a Creative Commons 
    Attribution 4.0 International License.
    License details:
    \url{http://creativecommons.org/licenses/by/4.0/}.
}

Data-to-Text generation (D2T) is defined as automatically generating natural language texts from non-linguistic inputs~\cite{reiter2000building}. 
Interest in this task has been driven by its applicability to specialized domains. For instance, D2T has been applied to generating weather reports~\cite{liang-etal-2009-learning}, restaurant descriptions~\cite{novikova2017e2e}, and video game dialogues~\cite{juraska2019viggo}. Recently, researchers have investigated D2T with more diverse domains to arrive at more generalizable text generation (such as works on \ldc~\cite{knight2017abstract} and \webnlg~\cite{gardent2017creating} datasets).

Traditional approaches to D2T follow a pipeline-based methodology, dividing the problem into several sub-problems~\cite{reiter2000building,gatt2018survey}. These include content selection (which information to include in the text), text structuring (the order in which to present the data), sentence aggregation (which information goes in individual sentences), lexicalization (finding the right words and phrases to express the data), referring expression generation (selecting the words and phrases to identify domain objects), and linguistic realization (combining all the generated words and phrases into well-formed sentences).

  In recent years, there has been a growing interest in going beyond pipeline-based approaches towards end-to-end (E-to-E) methods driven by recent advancements in deep learning~\cite{lebret-etal-2016-neural,novikova2017e2e,castro-ferreira-etal-2019-neural,duvsek2020evaluating}.
Such methods can be trained with (\textit{data,text}) tuples %
that can be efficiently collected at scale. In contrast, each step in pipeline-based approaches
requires its own setup and training data, such as semantic alignments between sections of text and components of the meaning representation (\mr). This makes them more costly and complex to develop and more prone to error propagation.

To date, end-to-end D2T %
has faced two main challenges: (1) \textbf{generalization} to unseen domains and (2) maintaining \textbf{semantic fidelity} to accurately convey the source data. In a recent comparative study, \newcite{castro-ferreira-etal-2019-neural} found that, compared to the best pipeline-based system, E-to-E  approaches based on GRU and Transformer architectures scored more than
35 BLEU points lower on unseen domains from the \webnlg dataset, and scored worst for semantic accuracy.

To address these challenges, we introduce \system, an E-to-E, domain-independent D2T system that makes no assumptions about the generated text's domain or the \mr%
's structure. 
\system leverages a pretrained language model and fine-grained state embeddings to achieve strong generalization. It also employs a weakly-supervised Semantic Fidelity Classifier (SFC) to detect and avoid generation errors (such as hallucination and omission). %
We also leverage this classifier to assess outputs from any D2T system, overcoming the limitations of existing heuristic %
methods for detecting semantic errors.

In this work, we deliver four main contributions across four major D2T datasets from various domains and MRs. %

\begin{itemize}
    \item 
    We show that \system pushes the state of the art on automated metrics by significant margins, ranging from 1.2 to 5.9 BLEU points, compared to the best existing pipeline and E-to-E techniques.
    \item
    With a crowdsourcing experiment, we demonstrate that \system generates text with significantly better fluency than existing works. On two datasets, our texts are even judged to be better, on average, than human-written references.
    \item We show that \system improves the semantic accuracy of generated texts, with margins ranging from 5.3\% to 40\% as assessed by crowdsourcing workers.
    \item With expert annotations, we further show that our model-based semantic accuracy metric is 4.2\% to 14.2\% more accurate in detecting semantic errors than existing heuristic-based approaches.
    
\end{itemize}

We open-source the \system code at \url{https://github.com/amazon-research}.

\section{Related Work}

\textbf{Pipeline vs. End-to-End Approaches}:
Within the pipeline-based paradigm, several studies have illustrated that breaking the D2T problem into sub-problems improves overall performance. \newcite{moryossef-etal-2019-step} showed that separating planning from realization helps achieve better semantic faithfulness compared to an E-to-E neural approach on the \webnlg dataset. 
\newcite{castro-ferreira-etal-2019-neural} conducted a comparative study across a variety of E-to-E and pipeline approaches with \webnlg, concluding that the latter are significantly better at generalizing to unseen domains.
However, so far the E-to-E approaches in these studies have been trained from scratch on the task dataset. 
Our work investigates whether using a pretrained model with strong language %
generation capabilities raises the performance of E-to-E models. %

\noindent\textbf{Structured Representations of the Data:}
Another thread of research focuses on %
better encoders for meaning representation languages, exploiting their structural properties. This is particularly relevant to AMR ~\cite{damonte-cohen-2019-structural,ribeiro2019enhancing,zhu-etal-2019-modeling,guo-etal-2019-densely}. \newcite{damonte-cohen-2019-structural} showed that replacing sequential encoders with a graph encoder improves text quality as measured by BLEU and METEOR scores. \newcite{zhu-etal-2019-modeling} proposed using self-attention to better model %
indirectly connected AMR components. 
In this work, we are the first to design a system that achieves a strong performance across different data structures, ranging from slot-value pairs to graph-based \mr. We also show that such a system can deliver significant gains compared to existing specialized systems.

\noindent\textbf{Semantic Fidelity Guarantees:}
To improve semantic fidelity (how accurately the generated text conveys the meaning) in E-to-E architectures, one approach has been to train reverse ``Text-to-Data'' models~\cite{Chisholm2017,agarwal-etal-2018-char2char}. 
Another approach by \newcite{kedzie-mckeown-2019-good} used data augmentation and a reliable \mr parser to reduce semantic errors in the generated text. \newcite{nie-etal-2019-simple} focused on fixing training data errors via an iterative data refinement technique using a language understanding module. 
\newcite{nie-etal-2018-operation} tackled the specific case where symbolic operations (e.g. numerical comparisons) are needed, augmenting the encoded input by pre-calculating these inferrable facts. 
\newcite{shen-etal-2019-pragmatically} used techniques from computational pragmatics and modeled the generation task as a game between speakers and listeners.
Despite following the generation-reranking paradigm explored previously in the data-to-text domain~\cite{agarwal-etal-2018-char2char,moryossef2019improving,Dusek2019}, and in other domains including machine translation~\cite{shen2004discriminative}, dialogue generation~\cite{wen2015stochastic}, and ASR~\cite{morbini2012reranking}, our work has several distinctive aspects compared to previous works. First, we do not make extra assumptions, such as availability of precise \mr parsers. 
Second, our system provides improvements even when the data is not the root cause of semantic errors. Third, we go beyond \emph{encouraging} the model to avoid semantically inconsistent outputs: we aim to also \textit{detect} with high probability when the generated text still contains such errors. 

For industrial NLG applications, including in healthcare \cite{pauws2019making} or news \cite{leppanen2017data}, identifying individual generations that are inaccurate is vital for the system to be useful in practice \cite{smiley-etal-2017-say}.
This error detection task has commonly relied on handwritten mappings from data values to potential realizations. Such rules were used to compute a Slot Error Rate (SER) metric~\cite{Dusek2019,juraska2019viggo,moryossef2019improving}. For instance, \newcite{Dusek2019} use SER for reranking beam elements during decoding in an attention-based sequence-to-sequence model on the \cleanedee dataset. \newcite{juraska2019viggo} used the approach similarly with a transformer model on the \viggo dataset. This technique %
is difficult to scale to new domains or languages, and struggles when the \mr is not dominated by values that occur verbatim in the text (e.g. named entities). 
We aim to tackle that with our model-based semantic fidelity classifier.

\section{Problem Description}

The D2T task is formally defined as generating text $T$ from data $D$ that is encoded via a meaning representation \mr. %
We assume that content selection is done prior to %
the D2T task, an assumption also made in the datasets we use. Therefore, the text $T$ should have \textit{semantic fidelity} by conveying all the input data, and only the input data. 

\subsection{Datasets}

We selected the major datasets that satisfy the task definition above. Each dataset consists of ($D$,$T$) pairs with texts in English. The following describes each dataset and our preprocessing/linearization, 
including special tokens added  (highlighted in $<$\textbf{bold}$>$ below) 
to guide our models. We provide the datasets' statistics in Table~\ref{tab:dataset_text_stats} of the appendix, and we show examples from them in Figure~\ref{fig:dataset_examples}.

\paragraph{\webnlg:} In \webnlg, $D$ is a set of 1-7 DBpedia triples which $T$ verbalizes ~\cite{gardent2017creating}. %
The test data spans 15 %
domains, 10 of which are seen in training. We linearize by concatenating triples, adding special tokens for `subject', `predicate', and `object', and converting strings %
to sentence-case. For fair comparison with the state of the art, we use v1.4 from~\newcite{castro-ferreira-etal-2018-enriching}.

\begin{figure}[b]
\begin{minipage}[c]{0.63\textwidth}
\begin{lstlisting}
WebNLG
D= Aarhus | leaderName | Jacob_Bundsgaard
Linearized D= <subject> Aarhus <predicate>  leader name 
              <object> Jacob Bundsgaard
T= The leader of Aarhus is Jacob Bundsgaard.
\end{lstlisting} 
\begin{lstlisting}
Cleaned E2E
D= name[Zizzi], eatType[coffee shop], area[riverside]
Linearized D= <name> name=[Zizzi]; <area> area=[riverside]
              <eatType> eatType=[coffee shop]; 
T= You can find a coffee shop named Zizzi in the riverside area.
\end{lstlisting}
\begin{lstlisting}
ViGGO
D= request( developer[EA Canada], specifier[favorite])
Linearized D= <request> request (<developer> developer: [EA Canada], 
              <specifier> specifier: [favorite] <request>)
T= What's your favorite game that EA Canada has made?
\end{lstlisting}
\end{minipage}
\begin{minipage}[c]{0.37\textwidth}
\begin{lstlisting}
LDC2017T10
D= (r / respond-01
 :ARG0 (c / country :wiki ``United_States'' 
   :name (n / name :op1 ``United" 
   :op2 ``States''))
 :ARG1 (d / develop-01
   :mod (t / that))
 :ARG2 (c2 / condemn-01
   :manner (s / swift)))
Linearized D= (respond <:ARG0> 
 (country <:name> (United States)) 
 <:ARG1> (develop <:mod> (that)) 
 <:ARG2> (condemn <:manner> (swift)))
T= The United States responded to that 
   development with swift condemnation.
\end{lstlisting}
\end{minipage}
\caption{Examples from each of the datasets}
\label{fig:dataset_examples}
\end{figure}

\paragraph{\ldc:}
In the \ldc dataset~\cite{knight2017abstract}, $D$ is an Abstract Meaning Representation (AMR) graph representing ``who is doing what to whom'' for each sentence in $T$.  The texts include broadcast news and weblogs. We use the preprocessing script from ~\newcite{ribeiro2019enhancing}, without lowercasing. We merge leaves that correspond to one entity (e.g. ``\textit{United States}'' below). Each role specifier %
is replaced with a special token.

\paragraph{\cleanedee:} The \cleanedee dataset introduced in~\cite{Dusek2019} is an automatically cleaned version of the original E2E dataset~\cite{novikova2017e2e}, aiming to eliminate omissions and hallucinations in the human text by fixing the corresponding \mr. Each \mr consists of 3-8 slot-value pairs in the restaurant domain. We preprocess $D$ by adding special tokens before each slot type.

\paragraph{\viggo:} In \viggo~\cite{juraska2019viggo}, $D$ is a meaning representation with one of 9 dialogue acts (e.g. \textit{give\_opinion, suggest}) and 1-8 slot-value pairs from 14 video game attributes (e.g. NAME, GENRES). Each $T$ is an utterance representing a dialogue turn. %
We add special tokens at the start and end, representing the dialog act, and %
before each slot type.

As we illustrate in Table~\ref{tab:dataset_text_stats}, the datasets vary widely. \ldc dataset is not bounded to specific domains. Hence, although the AMR format closely describes the text, it is non-trivial to generalize from the training to test data. \webnlg covers a wide, but restricted set of domains, only a subset of which are present in the training data.
However it has high lexical diversity. The number of unique words in the test set of \webnlg is 7253 (63\% of them capitalized), compared to 5533 (22\% capitalized) for \ldc, 2014 (33\% capitalized) for \viggo, and 1966 (29\% capitalized) for \cleanedee. Measured with the New Dale–Chall readability score~\cite{dale1948formula}, \ldc has the highest difficulty score (6.49) compared to 1.03, 0.85, and 1.02 for the \webnlg, \cleanedee, and \viggo datasets respectively. 
In terms of quality, \viggo was designed with the goal of perfect semantic fidelity, and \cleanedee was heavily filtered from the original dataset to achieve that. On the other hand, the versions we use of the other datasets have not undergone such filtering.

\begin{table}[t]
\begin{center}
\resizebox{1\textwidth}{!}{%
\begin{tabular}{lllllll} 
 \textbf{Dataset} & \textbf{Train Size} & \textbf{Validation Size} & \textbf{Test Size} & \textbf{Unique Words} & \textbf{\% Capitalized} & \textbf{Dale-Chall} \\
 \toprule
\href{https://github.com/tuetschek/e2e-cleaning/tree/master/cleaned-data}{\cleanedee} & 33525 & 4299  & 4693 & 1966 & 29\% & 0.85 \\
\href{https://catalog.ldc.upenn.edu/LDC2017T10}{\ldc} & 36521 &1368  & 1371 & 5533 & 22\% & 6.49 \\
 \href{https://nlds.soe.ucsc.edu/viggo}{\viggo} & 5103 & 714 & 1083 & 2014 & 33\% & 1.02 \\
\href{https://github.com/ThiagoCF05/webnlg/tree/master/data/v1.4/en}{\webnlg} & 18081 & 2260 & 4928 & 7253 & 63\% & 1.03
\end{tabular}
}
\caption{Dataset text statistics \label{tab:dataset_text_stats}}
\end{center}
\end{table}

\section{\system Architecture}
\label{sec:sys}

We designed \system to be highly generic in order to tackle diverse meaning representations and allow D2T generators to be built for new datasets
with minimal work beyond data preprocessing.
At a high-level, our text generation system takes a 2-stage approach: \textit{generation} and \textit{reranking}.
First, we fine-tune a pretrained language model on the D2T task using the task's training data. Next, we build a specialized semantic fidelity classifier trained on an automatically-generated task-specific corpus. Using these models, we construct a customized beam-search decoder that ranks candidates based on the probabilities from the language model, and, at its final stage, reranks them based on the classifier's labels.

\subsection{Data-to-Text Model Fine-tuning}
\label{sec:finetune}

\begin{figure*}[h]
    \centering
    \includegraphics[width=\textwidth]{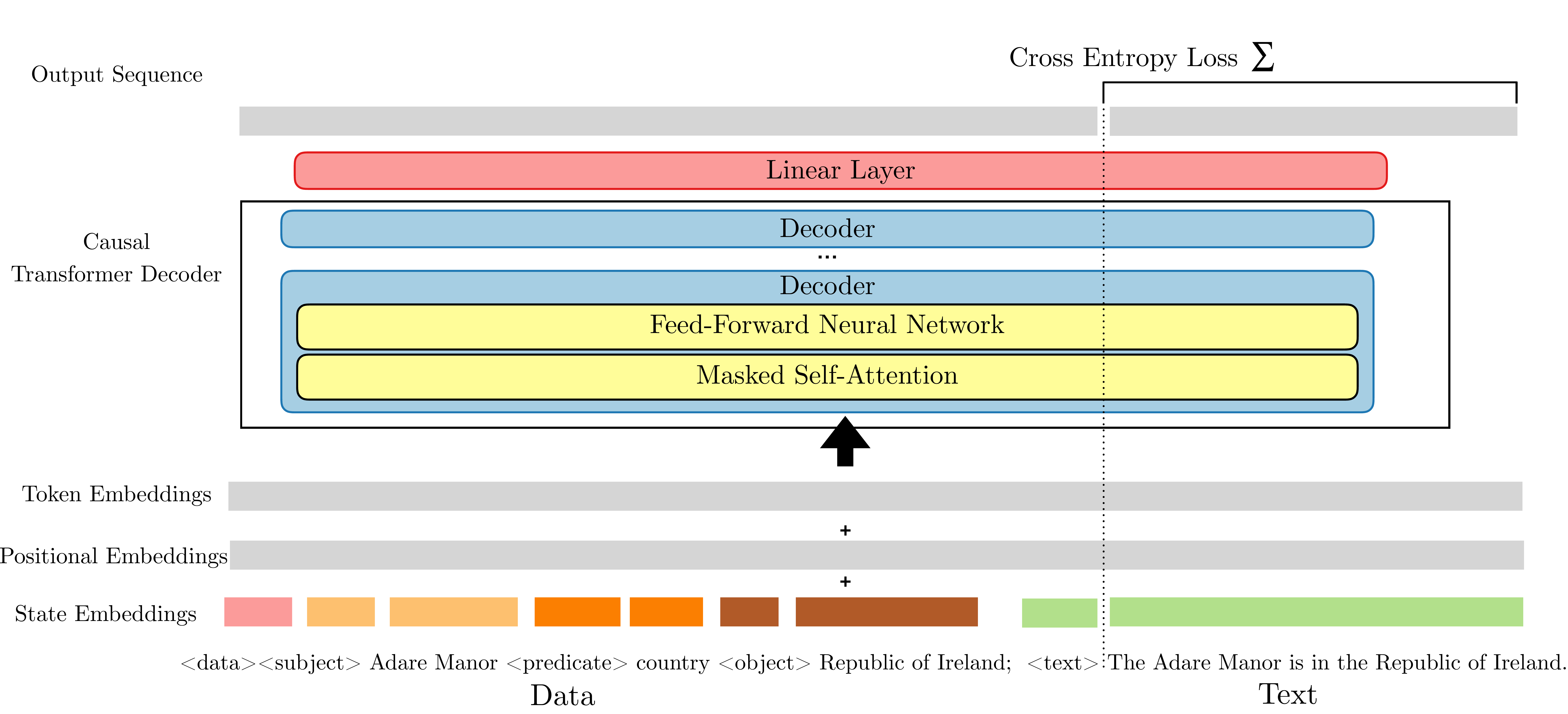}
    \caption{Data-to-text language model fine-tuning setup}
    \label{fig:model_diagram}
    \vspace{-0.5\baselineskip}
\end{figure*}

The %
fine-tuned Data-to-Text Language Model (\dtlm) builds on the pretrained OpenAI GPT-2 model~\cite{radford2019language}, a multi-layer, autoregressive language model. Each layer is a transformer decoder block~\cite{vaswani_transformer} of masked multi-headed attention and a fully connected layer. 
We provide a full model diagram in Figure~\ref{fig:model_diagram}.

\textbf{Inputs:}
The input sequence is the data $D$ concatenated with the text $T$: $(\specialtoken{data} \{D\} \specialtoken{text} \{T\})$.
The special tokens \specialtoken{data} and \specialtoken{text} %
are appended to GPT-2's original vocabulary; their embeddings are learnt during fine-tuning. 
In addition, we append to the vocabulary the \mr-dependent %
special tokens described above. %
After tokenization, we get a sequence $S$ of subword tokens, which are encoded to point to vocabulary indices: $S = (\specialtoken{data}, d_1,\ldots d_k, \specialtoken{text}, t_1, \ldots t_m) =  (s_0,\ldots s_n).$

One interesting feature of GPT-2 is its use of Byte-Pair Encoding (BPE)~\cite{sennrich-etal-2016-neural} on bytes instead of unicode characters. Hence, with a modestly-sized subword vocabulary of around 50K, it can encode any input text and score any output sequence, without suffering from unknown tokens. This is beneficial for our task where named entities are common. 

GPT-2 additionally expects positional encodings to help capture the input tokens' order.
Our core addition to the model is a third type of input: fine-grained state embeddings. These are analogous to the ``Segment Embeddings'', introduced in BERT~\cite{devlin-etal-2019-bert} to distinguish between sentence pairs in the next sentence prediction task. 
However, in our case, the state is defined at a more fine-grained level to give the model a hint on the type of the data being handled. 
The state vector for $S$ is a vector of tokens with size $|S|$, with each token ID indicating the type of $s_i$. Our strategy is to decide the state based on the special tokens we inserted in the data processing stage. We use the following rule: the state token ID of any token $s_i$ is the ID of the last special token preceding it (i.e. in the range ($s_0 \ldots s_i$) inclusively).

\textbf{Training:} The input embeddings, positional embeddings, and state embeddings are summed together and fed to the first GPT-2 layer. The last GPT-2 layer output is then normalized using ``LayerNorm''~\cite{ba2016layer}
before passing it to a linear layer added on top. The weights of the latter are tied to the input embeddings. 
Finally, a softmax is applied to the linear layer's output to generate probability distributions of the output tokens.
Our training objective is a language modeling one where we aim to find the set of weights $\theta$ that minimize the cross-entropy loss $\ell = \sum_{i=|D|+2}^{|S|}{ \log P_{\theta}(s_i| s_0, \ldots s_{i-1})}$.

Note that, since our task is to generate text given the data, the cross-entropy loss is computed for the text following the input data. We mask the data component in the loss above and sum the loss from index $|D|+2$ (i.e., after the \specialtoken{text} token). We use \textit{AdamW} as an optimizer~\cite{loshchilov2018decoupled}.

\subsection{Semantic Fidelity Classifier}
The %
Semantic Fidelity Classifier (\sfc) provides an additional assessment of %
how accurately the generated text reflects the input data.
A text is deemed to possess semantic fidelity if it accurately conveys all of the input data without omitting any nor adding additional data. %
Our approach draws parallels between this task %
and natural language inference (NLI) tasks, where the goal is to determine whether a ``hypothesis'' is true, false, or undetermined given a ``premise''. Similarly, in semantic fidelity classification, we aim to determine if the text is ``accurate'' or contains some ``omission'', ``repetition'', ``hallucination'', or ``value errors'' given the data. 
We cast the problem as a sentence-pair classification task for the (\textit{Data, Text}) pairs, using RoBERTa~\cite{liu2019roberta} as a base encoder. This formulation has been successfully used for NLI problems before~\cite{devlin-etal-2019-bert}.

\textbf{Training Data Generation:}
The classifier's training data should consist of semantically faithful and semantically incorrect examples. %
We generate training data for the \sfc automatically from the training data of the main D2T task. We define a set of simple dataset-independent transformations that account for common errors in data-to-text generation. 
For each tuple ({$D_i,T_i$}) in the training data, we split the text $T_i$ into sentences, using the spaCy sentence tokenizer~\cite{spacy2}. We then generate a set of new tuples for the \sfc consisting of ($D_i, T_j, l$) for each of the labels $l$ below, generated as follows:

\begin{itemize}
    \item \textbf{Accurate}: This is the text $T_i$.
    \item \textbf{Omission}: Remove the shortest sentence in $T_i$ (to help detect subtle omissions). 
    \item \textbf{Repetition}: Take a random sentence in $T_i$ and insert it before another random sentence in $T_i$. 
    \item \textbf{Hallucination}: Select a random sentence from another training text $T_{j\neq i}$ and insert it before a random sentence in $T_i$.
    \item \textbf{Value Errors}: Select a random value $x$ that occurs verbatim in both $D_i$ and $T_i$, and replace $x$ in $T_i$ with a random other value from $D_i$. 
    For slot-based \mr (\cleanedee and \viggo), $x$ is selected from the slots' values. For graph-based \mr (\ldc), $x$ is selected from the graph's leaves. For RDF triples (\webnlg dataset), $x$ is chosen from the triples' subjects and objects. 
\end{itemize}

\begin{wrapfigure}{r}{0pt}
    \centering
    \includegraphics[width=0.3\linewidth]{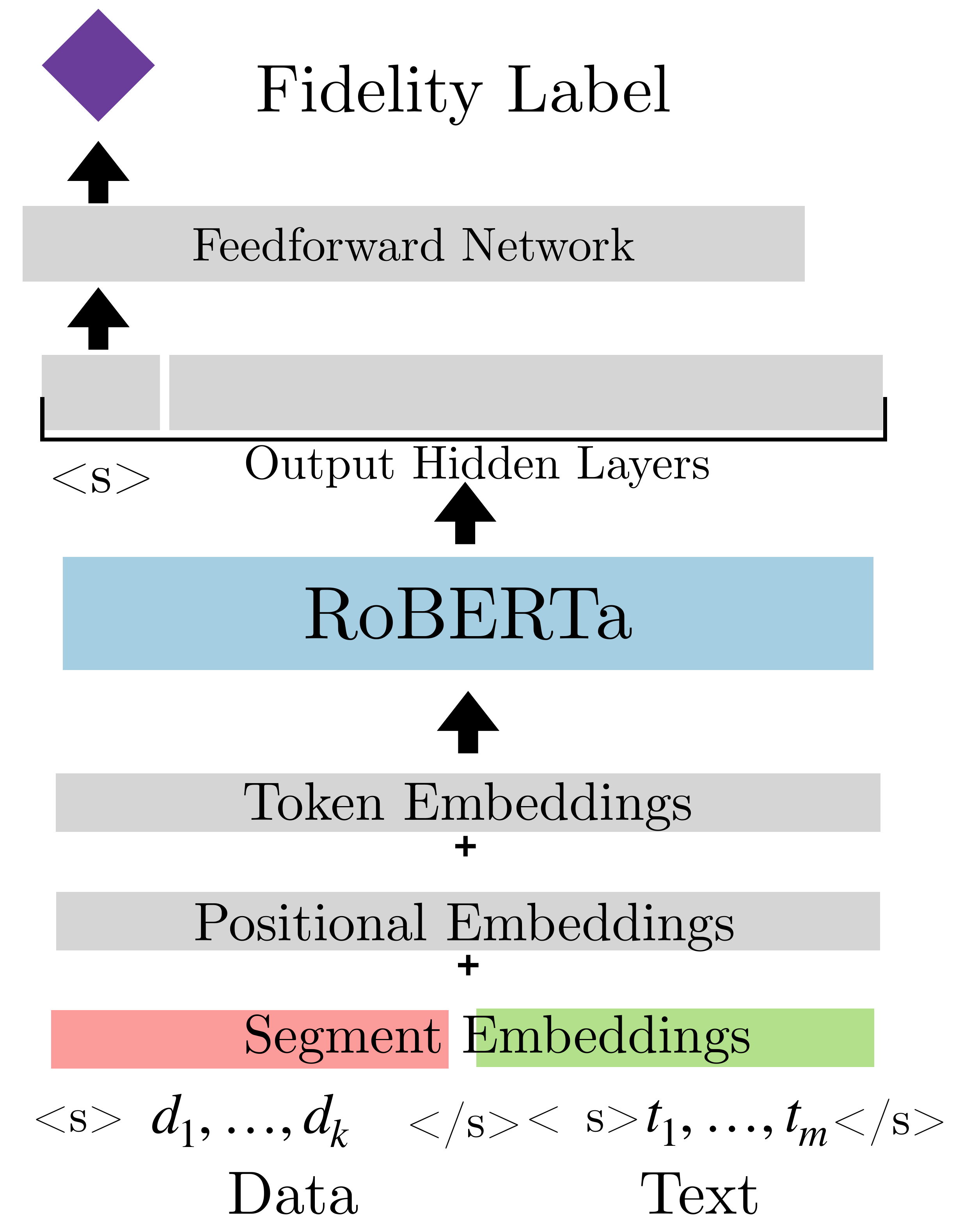}
    \caption{Semantic fidelity classifier setup}
    \label{fig:consistency_diagram}
\end{wrapfigure}

A related approach, with a different setup and modeling architecture, has been used before in the context of consistency in abstractive summarization~\cite{kryscinski2019evaluating}. There, weakly supervised models trained on domain-specific data have been shown to outperform supervised models trained on out-of-domain, human-annotated data.

\textbf{Model Input:} 
As shown in Figure~\ref{fig:consistency_diagram}, we concatenate the data and text tokens, adding the special start (\specialtoken{s}) and end (\specialtoken{/s}) tokens used during the training of RoBERTa. In addition to subword token embeddings, we add positional embeddings (representing the token position) and segment embeddings (representing data vs. text types). %

\textbf{Training:}
The 3 embeddings are summed element-wise to produce the input representation passed to RoBERTa's first encoder layer. Each layer subsequently applies a self-attention followed by a feed-forward network. 
We take the output hidden layer corresponding to the first token ($\specialtoken{s}$) and pass it through an additional single-layer neural network. The model is trained as a multi-class classifier %
with a cross-entropy loss as the objective and AdamW as the optimizer.

\subsection{Decoder} 
\label{sec:decoding}

Our decoding algorithm for the \dtlm is based on beam-search.
At each decoding step, items are ranked according to the score $R = \frac{1}{(i-(|D|+2))^{\alpha}}\displaystyle \prod_{|D|+2}^{i}{P(s_i|s_0 \ldots s_{i-1})}$. The score multiplies the conditional probabilities' product with a length normalization factor. Low-scoring candidates are dropped once the number of candidates exceeds the beam size.

Compared to traditional beam search, we do not aggregate probabilities from the start of the sequence, but from the start of the text component (index $|D| +2$). The length normalization is also adjusted to only account for the text component. 
We do this because we fine-tuned the \dtlm to generate text given data as context, not to generate the data itself. 
Hence, we remove the data tokens from the beam-scoring function. %
In our experiment, we use a value of $\alpha=0.75$.
At the end of the beam-search, we use the \sfc to rerank the \textit{complete} candidates (terminated with an end-of-sequence token) in the beam. The reranking metric uses the following binary score: $\mathbbm{1}_{SFC(D_i,T_i)=``accurate"}$.

Hence, we push the text $T_i$ to the top of the beam if our SFC labels the ($D_i,T_i$) tuple as ``\textit{accurate}''. We resolve ties using the original \dtlm scores.
An alternative strategy would be to apply reranking at each decoding stage, but we empirically found this strategy to have negligible accuracy gains %
while requiring a cost that grows with the text size. %
In addition to helping surface semantically accurate outputs, the \sfc labels can be used to assess whether the generated text is usable in practice. In our experiments, we compare this model-based approach to the heuristic approaches commonly used.

\section{Experiments}

For each dataset, we generate outputs from three versions of \system for our ablation studies.
 \systemNoFcNoFs simply relies on the \dtlm, with no \sfc-based reranking and a coarse-grained version of the state embeddings that contains only \specialtoken{data} and \specialtoken{text} tokens (as done by~\newcite{wolf_transfo}).
\systemNoFc adds the fine-grained state embeddings described in Section~\ref{sec:finetune} to \systemNoFcNoFs.
\systemFcPost adds the \sfc-based reranking. For the \sfc, we train the model using the RoBERTa-large model (355M parameters) on lower-cased text. On the synthetic test set generated, the %
classifier has a macro-averaged F1-score (across 5 classes) of 97\%, 97\%, 98\%, and 98\% for the \ldc, \webnlg, \cleanedee, and \viggo datasets respectively. We use the models bundled within the HuggingFace Transformers library~\cite{Wolf2019HuggingFacesTS}. The \dtlm uses the GPT-2-Medium model (with 345M-parameters) as its base model. The beam search width during decoding is 5. Training was performed on a single machine (Amazon AWS p3.8xlarge). During inference, text from DataTuner is generated at an average rate of 11.8 tokens per second on NVIDIA Tesla K80 GPUs.

We evaluate each variant's %
outputs with automated metrics and crowdsourced fluency and fidelity evaluation. We also quantify the efficacy of our semantic fidelity classifier with expert-annotations.
We compare against the state of the art systems on each dataset, selected based on BLEU scores. 
These are a graph-optimized Transformer Seq2seq model by \newcite{zhu-etal-2019-modeling} for \ldc, Transformer-based models (pipeline and E2E) by \newcite{castro-ferreira-etal-2019-neural} for \webnlg, an LSTM Seq2seq model with a rule-based reranker by \newcite{Dusek2019} for \cleanedee, and a Transformer Seq2seq model by \newcite{juraska2019viggo} for \viggo. %
The supplementary material contains the outputs from our system variants and the main training hyperparameters (obtained via manual tuning).

\subsection{Automated Evaluation}

For each test set, we compute BLEU (B)~\cite{papineni2002bleu}, which measures n-gram precision, METEOR (M) ~\cite{lavie2007meteor}, which is based on the harmonic mean of the unigram precision and recall while accounting for stem and synonymy matching, ROUGE$_L$ (R)~\cite{lin2004rouge}, which calculates recall for the longest common subsequence, and CIDEr (C)~\cite{vedantam2015cider}, which is based on TF-IDF scoring of n-grams. 
We used the official evaluation scripts of the E2E challenge\footnote{\url{https://github.com/tuetschek/e2e-metrics}}. 
Table~\ref{tab:automated_metrics} compares the results generated by \system variants against the state of the art.%

\textbf{Improvements from the \dtlm alone}: Comparing %
the simple \systemNoFcNoFs model to the state of the art, we find that it already improves %
the BLEU score across 2 %
datasets and the METEOR score across 3 datasets. 
This indicates that the \dtlm component of \system is itself contributing to achieving an end-to-end state-of-the-art system that needs no delexicalization or \mr-specific encoding.

\textbf{Fine-grained state embeddings matter:} Across the \numData datasets, adding fine-grained state embeddings boosts %
performance on these metrics, with improvements ranging from 0.3 (on \cleanedee) to 2.0 BLEU points (on \viggo).

\textbf{\sfc effect on automated metrics:} 
Several studies highlight shortcomings of automated metrics for evaluating semantic adequacy~\cite{novikova-etal-2017-need,shimorina2018human}. In this vein, compared to our \systemNoFc model, we observe slight additional 
boosts from introducing the \sfc classifier in the \systemFcPost variant. Interestingly, \systemFcPost always has the highest %
METEOR score, which was the only metric found by~\newcite{shimorina2018human} to correlate with semantic adequacy.

\begin{wraptable}[21]{r}{0.67\linewidth}
\resizebox{0.8\linewidth}{!}{%
\begin{tabular}{llcccc}
\toprule
\textbf{D}                   & \textbf{Model}                           & \textbf{B}            & \textbf{M}                     & \textbf{R}           & \textbf{C}\\
\midrule  
\vertmulticell{6}{\ldc}            
                                   &  \systemFcPost                              & \textbf{37.7}                  & \textbf{38.9}                  & \textbf{65.1}                &  \textbf{3.9}                  \\
                                   &  \systemNoFc                              & 37.2                  & 38.4                  & 65.0                  & \textbf{3.9}                 \\
                                   &  \systemNoFcNoFs                              & 35.6                  & 37.3                  & 64.4                  & 3.8                  \\

                                   & \newcite{zhu-etal-2019-modeling}           & 31.8                  & 36.4                    &  -                   &  -    \\
                                   & \newcite{guo-etal-2019-densely}             & 30.4                  &  -                      &  -                    &  -   \\
                                   & \newcite{ribeiro2019enhancing}              & 27.9                  & 33.2                    &  -                    & -   \\
\midrule       
\vertmulticell{6}{\webnlg}         

                                   &  \systemFcPost                              & 52.4                  & \textbf{42.4}                  & \textbf{66.0}                  & \textbf{3.7}                  \\
                                   &  \systemNoFc                              & \textbf{52.9}                  & 41.9                  & 65.9                  & \textbf{3.7}                  \\
                                   &  \systemNoFcNoFs                              & 51.6                  & 40.6                  & 64.9                  & 3.6                  \\

                                   & \newcite{castro-ferreira-etal-2019-neural} Pipe.  & 51.7                  &  32.0                   &    -                  &   -    \\
                                  & \newcite{castro-ferreira-etal-2019-neural} E2E  & 33.5                  &  25.0                   &  -                    &  -     \\

                                    & \newcite{moryossef-etal-2019-step} Pipe.       & 47.4                  &  39.1                   & 63.1                     & 2.7      \\
\midrule       
\vertmulticell{5}{\cleanedee}     
                                   &  \systemFcPost                              & \textbf{43.6}                  & \textbf{39.0}                  & 57.5                  & \textbf{2.0}                  \\
                                   &  \systemNoFc                              & \textbf{43.6}                  & \textbf{39.0}                  & 57.5                  & \textbf{2.0}                  \\
                                   &  \systemNoFcNoFs                              & 43.3                  & 38.9                  & \textbf{57.6}                  & \textbf{2.0 }                 \\

                                   &  \newcite{Dusek2019}  (TGen+)              &  40.5               & 37.6                   &   56.0                      &   1.8    \\
\\
\midrule       
\vertmulticell{4}{\viggo}          

                                   &  \systemFcPost                              & \textbf{53.6}                  & \textbf{39.4}                  & \textbf{64.0}                  & \textbf{2.7}                  \\
                                   &  \systemNoFc                              & 53.4                  & 39.1                  & 63.8                  & \textbf{2.7}                  \\
                                   &  \systemNoFcNoFs                              & 51.4                  & 38.9                  & 62.7                  & 2.5                  \\

                                   &  \newcite{juraska2019viggo}                 &  52.1               & 39.1                    &   63.8                  &   2.5    \\
\midrule                
\end{tabular}
}
\caption{Evaluation with automated metrics. }\label{tab:automated_metrics}
\end{wraptable}

\textbf{Largest boost on the most complex text:} \system had the biggest improvement, 5.9 BLEU points, on the \ldc dataset.  This is interesting given that (1) the text in \ldc is typically long with more complex sentence structures 
and that (2) the baseline systems targeting AMR-to-text~\cite{zhu-etal-2019-modeling,guo-etal-2019-densely,ribeiro2019enhancing} built more sophisticated architectures compared to other datasets (e.g. \viggo and \cleanedee). 
This illustrates our system's ability to work across a spectrum of data representations and text complexity.

\subsection{Human Evaluation of Fluency}
\label{sec:fluency}

We conduct human evaluation of fluency
for 150 examples sampled at random from each dataset. 
We use Amazon's MTurk to ask crowd workers how fluent a text is on a 7-point Likert scale using sliders, where ``high fluency'' is defined as ``grammatical, natural, and could have been produced by a skilled native speaker''.  Following findings from \newcite{novikova2018rankME} and \newcite{inlg2019evaluation} for acquiring more consistent human ratings, texts generated for the same meaning representation by different systems are presented together in a single task for annotators to score them relative to each other. We include the human-written text, and randomize the texts' order. 
For fair comparison, we lower-case our generated texts for the \ldc to match the outputs of \newcite{zhu-etal-2019-modeling}. We also detokenize outputs from that work to avoid these biasing the workers. %
We restrict to US-based annotators who completed $>$500 tasks, out of which more than 97\% had been accepted.

\textbf{Improvement on the state of the art:} As shown in the column ``Flu.'' of Table~\ref{tab:human_eval}, compared to the human baseline, our \systemFcPost model improves the fluency on all four datasets compared to the state of the art systems with statistically significant margins ($p<0.05$). For computing significance, we use the pairwise Wilcoxon signed-rank test~\cite{wilcoxon1992individual} with the null hypothesis that the fluency values for each pair of systems come from the same distribution.
For \ldc, where \systemFcPost had the largest gap in BLEU score (+5.9), we observe the widest fluency improvement (+0.82) compared to~\newcite{zhu-etal-2019-modeling}.
Interestingly, despite the fact that \systemFcPost scored 0.7 higher on BLEU compared to the pipeline approach in~\cite{castro-ferreira-etal-2019-neural} for \webnlg, the difference in fluency is 0.69, which is relatively large. We conjecture that this originates from two main sources. First, semantic errors might be perceived by annotators as breaking the fluency. For example, one text contained the phrase ``\textit{has a runway length of Shehbaz Sharif}''. Second, the pipeline approach had a sizeable portion of non-realized outputs (e.g. ``PATIENT-1 is made with PATIENT-1 and PATIENT-2.''), which were annotated as non-fluent. 
On the closed-domain datasets (\viggo and \cleanedee), the fluency margins are smaller while still statistically significant. This is expected as these datasets have a narrow set of sentence formulations that are easier to learn.

\textbf{Improvement on the human baseline}: Surprisingly, we find that \systemFcPost received a higher overall average fluency score on 3 %
datasets compared to the human baseline. This difference is statistically significant in both \cleanedee and \viggo, with the largest difference being 1.04 points in \cleanedee. Investigating, we found several low-scored texts had an informal style and problems in sentence construction. One example contained ``\textit{It serves Chinese food for less.}'' One explanation could be that, once fine-tuned on a large enough dataset, our models have less tendency to %
deviate from common formulations that annotators prefer.

\begin{wraptable}[28]{t}{0.65\linewidth}
\begin{center}
\resizebox{1\linewidth}{!}{%
\footnotesize
\begin{tabular}{>{\centering}p{0.01\linewidth}>{}p{0.37\linewidth}>{\centering}p{0.08\linewidth}>{\centering}p{0.08\linewidth}>{\centering}p{0.08\linewidth}>{\centering}p{0.05\linewidth}>{\centering}p{0.08\linewidth}>{\centering\arraybackslash}p{0.02\linewidth}}
\toprule
\textbf{D}            & \textbf{Model}           &   \textbf{Flu.}    &   \textbf{Fid.}    &   \textbf{DSA}  &   \textbf{\qdsa}      &   \textbf{HSA}      & \textbf{\qhsa} \\ %
\midrule
\vertmulticell{5}{\ldc}
                                 & \systemFcPost                             & $4.79^{s,h}$ & $78.0^{s}$    &  $81.8^{s,h}$   & \hormulticell{5}{90.8}       & $-^{}$    &   \hormulticell{5}{-}        \\
                                 & \systemNoFc                               & $4.87^{s}$   & $77.3^{s}$  &  $70.5^{s,h}$   &           & $-^{}$    &              \\
                                 & \systemNoFcNoFs                           & $4.78^{s,h}$ & $68.7$    &  $65.8^{s,h}$   &           & $-^{}$    &              \\
                                 & \newcite{zhu-etal-2019-modeling}            & $3.97^{h}$   & $58.7$  &  $62.4^{h}$   &           & $-^{}$    &              \\
                                 & Human                                     & $5.05^{}$    & $-^{}$ &  $93.1^{}$   &           & $-^{}$    &              \\

\midrule
\vertmulticell{5}{\webnlg}
                                 & \systemFcPost                             & $5.23^{s}$  & $84.0^{s}$   &  $91.7^{s,h}$   & \hormulticell{5}{\textbf{87.5}}       & $58.1^{s,h}$    &   \hormulticell{5}{73.3}        \\
                                 & \systemNoFc                               & $5.20^{s}$  & $80.0^{s}$   &  $81.4^{s,h}$   &           & $54.1^{s,h}$    &              \\
                                 & \systemNoFcNoFs                           & $5.27^{s}$  & $72.0^{s}$   &  $73.6^{s,h}$   &           & $43.6^{s,h}$    &              \\
                                 & \newcite{castro-ferreira-etal-2019-neural}  & $4.54^{h}$  & $44.0$   &  $50.5^{h}$   &           & $33.7^{h}$    &              \\
                                 & Human                                     & $5.21^{}$   & $-^{}$  &  $94.7^{}$   &           & $41.2^{}$    &              \\

\midrule
\vertmulticell{5}{\cleanedee}
                                 & \systemFcPost                             & $5.46^{s,h}$ & $92.0^{s}$    &  $99.3^{s,h}$   & \hormulticell{5}{\textbf{89.2}}       & $97.3^{s,h}$    &   \hormulticell{5}{75.0}        \\
                                 & \systemNoFc                               & $5.46^{s,h}$ & $90.7^{s}$    &  $99.0^{h}$   &           & $97.1^{s,h}$    &              \\
                                 & \systemNoFcNoFs                           & $5.45^{s,h}$ & $88.7$    &  $98.9^{h}$   &           & $97.1^{s,h}$    &              \\
                                 & \newcite{Dusek2019} TGen+                   & $5.23^{h}$   & $86.0$  &  $98.9^{h}$   &           & $98.0^{h}$    &              \\
                                 & Human                                     & $4.42^{}$    & $-^{}$ &  $99.9^{}$   &           & $100.0^{}$    &              \\

\midrule
\vertmulticell{5}{\viggo}
                                 & \systemFcPost                             & $5.77^{s,h}$  & $86.0$    &  $97.2^{s}$   & \hormulticell{5}{92.5}       & $74.5^{s,h}$    &   \hormulticell{5}{88.3}        \\
                                 & \systemNoFc                               & $5.76^{s,h}$  & $82.7$    &  $92.8^{h}$   &           & $82.3^{s,h}$    &              \\
                                 & \systemNoFcNoFs                           & $5.60^{h}$    & $81.3$  &  $91.7^{h}$   &           & $82.5^{s,h}$    &              \\
                                 & \newcite{juraska2019viggo}                  & $5.58^{h}$    & $80.7$  &  $92.8^{h}$   &           & $90.9^{}$    &              \\
                                 & Human                                     & $5.34^{}$     & $-^{}$ &  $97.1^{}$   &           & $91.9^{}$    &              \\

\midrule
\end{tabular} 
}
\caption{Crowdsourced evaluation of fluency (Flu.), crowdsourced evaluation of semantic fidelity (Fid.), \system Semantic Accuracy (DSA), Heuristic Semantic Accuracy (HSA), and quality measures \qdsa and \qhsa for DSA and HSA. Superscripts $s$ and $h$ indicate a statistically significant difference vs. the state of the art and the human baseline respectively. Bolded $Q_{D}$ values are higher with  statistical significance when compared to the corresponding $Q_{H}$ values.}\label{tab:human_eval}
\end{center}
\end{wraptable}

\subsection{Crowdsourced Evaluation of Fidelity}
We take a two-step approach for evaluating semantic fidelity with both crowdworkers and expert annotators.
We start with a crowdsourcing experiment involving the same 150 randomly sampled examples from each dataset used in the fluency evaluation. We use Amazon's MTurk (with the same restrictions on annotators) and present texts for the same \mr together. To avoid requiring non-expert annotators to understand the \mr, we present a human reference text against which to compare the system outputs. We ask annotators to make a choice whether each text is ``accurate'' or ``inaccurate'', i.e. whether it ``conveys all the factual information in the original text'' without any ``information missing, added or repeated''. We ask annotators to ignore grammar quality or style differences, provided that the overall meaning is the same. Three annotators complete each task, and we take the mode result for each text, assuming ``inaccurate'' in the event of a tie. 

\textbf{\system has higher semantic accuracy in human evaluation}
 The results presented in the ``Fid.'' column of Table~\ref{tab:human_eval} show that  \systemFcPost has a superior accuracy to all the other variants as well as the state of the art models. The differences to the state of the art models range from 5.3\% (\viggo) to 40\% (\webnlg) and are statistically significant on \webnlg, \ldc, and \cleanedee ($p<0.05$, as measured by McNemar's test~\cite{mcnemar1947note}).
 The trend among the \system variants also shows a clear impact of both the fine-grained state embeddings and the \sfc on boosting the overall accuracy of the generated text.

\subsection{Expert Evaluation of Fidelity}
Next, we assess the semantic fidelity with experts' annotations. We have two goals in this section: (1) comparing our model-based approach to heuristic-based approaches as automated methods of judging semantic accuracy, and (2) using this comparison outcome to illustrate that \system delivers higher semantic accuracy as measured by the better fidelity metric.

The baseline method uses heuristics to label each data-text tuple as accurate ($A_H$) or erroneous ($E_H$). For this, we use the heuristics by~\newcite{shimorina2018handling} for \webnlg, by~\newcite{juraska-etal-2018-deep} for \viggo, and by~\newcite{Dusek2019} for \cleanedee. We are not aware of heuristic-based scripts for \ldc. We compute Heuristic Semantic Accuracy (HSA) of a dataset as the fraction with the label $A_H$.
Our method uses the \sfc component in \system to assign accurate ($A_D$) or erroneous ($E_D$) labels to each data-text tuple. We compute \system Semantic Accuracy (DSA) as the fraction with the label $A_D$.
Both metrics are computed per system across each dataset.

To compare the quality of HSA and DSA as measures of semantic accuracy, we manually annotated a sample of data-text tuples. 
Since the vast majority of texts are expected to be accurate, especially on the cleaner datasets, we designed a sampling methodology to give a balanced representation of semantically accurate and inaccurate texts. 
To start, we sample 4 indices from the target dataset such that the human baseline outputs for these indices are labeled as: $\{(A_{H},E_{D}), (E_{H},A_{D}), (E_{H},E_{D}), (A_{H},A_{D})\}$.
We do the same with the state of art system and \systemFcPost outputs. We continue in a round-robin fashion until we get 24 indices per dataset. For \ldc dataset, we sample 24 indices in a similar fashion while ignoring the $A_H$ and $E_H$ labels.
Next, two authors were presented with the input meaning representation and the output texts generated by each system (in a random order) for the 24 sampled entries. 
The authors manually labeled the resulting 480 data-text tuples as accurate ($A_M$) or erroneous ($E_M$). Inter-annotator agreement measured with Cohen's Kappa was 0.81, indicating near-perfect agreement.
We use these labels to assess the quality \qdsa of the DSA metric as the percentage of cases where the manual label $A_M$ matches $A_D$. Similarly, we evaluate the quality \qhsa of the HSA metric as the percentage of cases where $A_M$ matches $A_H$. These percentages are aggregated across systems, obtaining 120 samples per dataset. We present these metrics in Table~\ref{tab:human_eval}.

\textbf{DSA provides higher quality semantic evaluation}: We notice that \qdsa is 4.2\% higher on \viggo and 14.2\% higher on both \cleanedee and \webnlg, compared to \qhsa. These differences are statistically significant ($p<0.05$) on \webnlg and \cleanedee, as measured by McNemar's test~\cite{mcnemar1947note} with the null hypothesis that the marginal probability for each outcome (\textit{accurate} or \textit{erroneous}) is the same for both algorithms. 

\textbf{HSA struggles with open domains}: The heuristic-based approach labeled only 41.2\% of the human references in \webnlg as accurate, 16.9\% lower than the score it assigned to our \systemFcPost. Since the latter was trained on human references, this difference is more likely to stem from shortcomings of this approach for assessing the semantics. Checking the data, we observed that humans tend to create more diverse formulations, such as converting \textit{United Kingdom} to \textit{UK}, which are easy to miss with heuristics. On the contrary, our DSA metric scored the human references higher.

\textbf{\systemFcPost delivers higher semantic accuracy}: 
Now that we have established that DSA is a better measure of semantic accuracy compared to HSA, we can see from Table~\ref{tab:human_eval} that, across all datasets, \systemFcPost significantly improves over the state of the art models as measured by the DSA metric (McNemar's test gives $p<0.05$). Compared to other \system variants, \systemFcPost adds between 0.3\% and 11.3\% improvements, corroborating the utility of the semantic fidelity classifier.
Finally, we note that, since the baseline models for \cleanedee and \viggo use the heuristics for reranking their outputs, they are expected to show higher HSA. However, what our manual annotations prove is that the HSA metric itself is of lower quality compared to the DSA metric.

\section{Conclusion}
We presented \system, an end-to-end data-to-text generation system equipped with a %
semantic fidelity classifier. \system records new state of the art results on four different datasets, with significant margins on automated metrics. We also show that our system has a clear fluency advantage over all the previous state of the art models. We further illustrate \system's strengths for delivering semantically accurate outputs.

\bibliographystyle{coling}
\bibliography{coling2020}

\clearpage

\section*{Appendix A. Examples of Generated Text}

In Table~\ref{tab:humanExamples}, we present examples of the outputs generated by \system and the state of the art models, alongside the human references.

\begin{table}[h]
\begin{center}
\resizebox{0.8\linewidth}{!}{%
\begin{tabular}{llp{10cm}}
\toprule
\textbf{D}     & \textbf{Model}       & \textbf{Examples}    \\
\midrule
\vertmulticell{7}{\cleanedee}  &    Human  & pub in riverside that also serves Italian food: The Vaults has got high prices, is not child-friendly and has an average rating. It is near Rainbow Vegetarian Café in riverside.\\
& \systemFcPost & The Vaults is an Italian pub in the riverside area near Rainbow Vegetarian Café. It has an average customer rating and a high price range. It is not child friendly. \\
& \newcite{Dusek2019} & The Vaults is an italian pub with an average customer rating. It is located in the riverside area near Rainbow Vegetarian Café. It is not children friendly and has a high price range. \\
& & \\
\vertmulticell{5}{\webnlg}  & Human &  Adirondack regional airport serves both Lake Placid and Saranac Lake, New York. The length of the runway at Asirondack regional airport is 2003.\\
 & \systemFcPost & Adirondack Regional Airport serves the cities of Lake Placid and Saranac Lake, New York and has a runway length of 2003.0. \\
   & \newcite{castro-ferreira-etal-2019-neural} & Adirondack Regional Airport serves the city of Lake Placid and Saranac Lake, New York and has a runway length of Shehbaz Sharif.\\
   \\
   \vertmulticell{5}{\ldc}  & Human &  the plan requires 8 precautionary steps before the order to shoot down a plane may be issued. \\
 & \systemFcPost &  the plan requires eight precautionary steps before the order to shoot down the plane can be issued.\\
   & \newcite{zhu-etal-2019-modeling} & the plan required 8 precaution steps before it can be issued to order shot down.\\
   \\
   \vertmulticell{5}{\viggo}  & Human &  Guitar Hero: Smash Hits was a very bad game. 2009 was a terrible year for gaming and I just can't stand the games released that year.  \\
 & \systemFcPost &  Guitar Hero: Smash Hits is a really bad game. 2009 was a really bad year for games \\
   & \newcite{juraska2019viggo} & Guitar Hero: Smash Hits is a very bad game, especially because it came out in 2009. \\
\midrule
\end{tabular} 
}
\caption{Examples of text generated by the different models}\label{tab:humanExamples}
\end{center}
\end{table}

\end{document}